\documentclass[11pt,a4paper]{article}
\usepackage[hyperref]{emnlp-ijcnlp-2019}
\usepackage{times}
\usepackage{latexsym}
\usepackage{tabularx}
\usepackage{amssymb}
\usepackage{amsmath}
\usepackage{enumitem}
\usepackage{booktabs}
\usepackage{url}
\usepackage{color} 
\usepackage{verbatim}
\usepackage{calc}
\usepackage{mathtools}
\usepackage[draft]{todonotes}
\usepackage{soul}
\usepackage{tabularx}
\usepackage[utf8]{inputenc}
\usepackage{pbox}
\usepackage{color, colortbl}
\definecolor{Gray}{gray}{0.9}
\definecolor{LightCyan}{rgb}{0.88,1,1}

\aclfinalcopy 

\title{You Shall Know a User by the Company It Keeps:\\
Dynamic Representations for Social Media Users in NLP}

\author{
	Marco Del Tredici \\
  University of Amsterdam \\
  {\tt m.deltredici@uva.nl} \\\And
  Diego Marcheggiani\thanks{\enspace Research conducted when the author was at the University of Amsterdam.}   \\
  Amazon \\
  {\tt marchegg@amazon.es} \\\AND
  Sabine Schulte im Walde \\
  University of Stuttgart \\
  {\tt schulte@ims.uni-stuttgart.de} \\\And
  Raquel Fern\'andez \\
  University of Amsterdam \\
  {\tt raquel.fernandez@uva.nl}}

\date{}

\begin{document}
\maketitle
\begin{abstract}
  Information about individuals can help to better understand what they say, particularly in social media where texts are short. Current approaches to modelling social media users pay attention to their social connections, but exploit this information in a static way, treating all connections uniformly. This ignores the fact, well known in sociolinguistics, that an individual may be part of several communities which are not equally relevant in all communicative situations. We present a  model based on Graph Attention Networks that captures this observation. It dynamically explores the social graph of a user, computes a user representation given the most relevant connections for a target task, and combines it with linguistic information to make a prediction. 
We apply our model to three different tasks, evaluate it against alternative models, and analyse the results extensively, showing that it significantly outperforms other current methods.
\end{abstract}

\section{Introduction}
\label{sec:introduction}

The idea that extra-linguistic information about speakers can help language understanding has recently gained traction in NLP. Several studies have successfully exploited social information to classify user-generated language in downstream tasks such as sentiment analysis~\cite{yang2017overcoming}, abusive speech identification~\cite{mishra2018author} and sarcasm detection~\cite{hazarika2018cascade,wallace2016modelling}. 
The underlying goal is to capture the sociological phenomenon of \textit{homophily} \cite{mcpherson2001birds} -- i.e., people's tendency to group together with others they share ideas, beliefs, and practices with -- and to exploit it jointly with linguistic information to obtain richer text representations.
In this paper, we advance this line of research. In particular, 
we address a common shortcoming in current models by using state-of-the-art graph neural networks to encode and  leverage homophily relations.

 
Most current models represent speakers as multidimensional vectors derived by aggregating information about all known social relations of a given individual in a uniform way. 
A major limit of this approach is that it does not take into account the well known sociolinguistic observation that speakers typically belong to several communities at once, in the sense of `communities of practice' \cite{eckert-mcconnellginet1992} denoting an aggregate of people defined by a common engagement, such as
supporters of a political party or fans of a TV show. Membership to these communities has different relevance in different situations. For example, consider an individual who is both part of a supporters group of the USA Democratic Party and of a given sports team. While membership to these two communities can be equally important to characterise the person in general terms, the former is much more relevant when
it comes to predicting whether linguistic content generated by this person expresses a certain stance towards president Trump. 
Current models fail to capture this context-dependent relevance.

In this work, we address this shortcoming, making the following contributions:

\begin{itemize}[leftmargin=11pt,itemsep=0pt, topsep=1pt]
\item We use Graph Attention Networks~\cite{velickovic2018graph} to design a model that dynamically explores the social relations of an individual, learns which of these relations are more relevant for the task at hand, and computes the vector representation of the target individual accordingly. This is then combined with linguistic information to perform text classification tasks.

\item To assess the generality and applicability of the model, we test it on three different tasks (sentiment analysis, stance detection, and hate speech detection) using three annotated Twitter datasets and evaluate its performance against commonly used models for user representation.

\item We show that exploiting social information leads to improvements in two tasks (stance and hate speech detection) and that our model significantly outperforms competing alternatives. 

\item We perform an extended error analysis, in which we show the robustness across tasks of user representations based on social graphs, and the superiority of dynamic representations over static ones.

\end{itemize}

\section{Related Work}
\label{sec:related_work}

Several strands of research have explored different social features to create user representations for NLP in social media. 
\newcite{hovy2015demographic} and \newcite{hovy2018improving} focus on demographic information (age and gender), while 
\newcite{bamman2014distributed} and \newcite{hovy2018capturing} exploit geographic location to account for regional variation. 
Demographic and geographic information, however, need to be made explicit by users and thus are often not available or not reliable. To address this drawback, other studies have aimed at extracting user information by just observing users' behaviour on social platforms. 
 
To tackle sarcasm detection on Reddit, \newcite{kolchinski2018representing} assign to each user a random embedding that is updated during the training phase, with the goal of learning individualised patterns of sarcasm usage.
\newcite{wallace2016modelling} and \newcite{hazarika2018cascade} address the same task, using ParagraphVector \cite{le2014distributed} to condense all the past comments/posts of a user into a low dimensional vector, which is taken to capture their interests and opinions. 
All these studies use the concatenation of author and post embeddings for the final prediction, showing that adding author information leads to significant improvements.



While the approaches discussed above consider users individually, a parallel line of work has focused on leveraging the social connections of users in Twitter data. 
This methodology  relies on creating a social graph where users are nodes connected to each other by their retweeting, mentioning, or following behaviour. 
Techniques such as Line \cite{tang2015line}, Node2Vec \cite{grover2016node2vec} or Graph Convolutional Networks \cite[GCNs,][]{kipf2017semi} are then used to learn low-dimensional embeddings for each user, which have been shown to be beneficial in different downstream tasks when combined with textual information.
For example, \newcite{mishra2018author} and \newcite{mishra2019abusive} use the concatenation strategy mentioned above for abusive language detection; \newcite{yang2016toward} optimise social and linguistic representations with two distinct scoring functions to perform entity linking; while \newcite{yang2017overcoming} use an ensemble learning setup for sentiment analysis, where the final prediction is given by the weighted combination of several classifiers, each exploring the social graph independently.

%


Methods like Line, Node2Vec and GCNs create user representations by aggregating the information coming from their connections in the social graph, without making any distinction among them. 
In contrast, we use Graph Attention Networks \cite[GATs,][]{velickovic2018graph}, a recent neural architecture that applies self attention to assign different relevance to different connections, and computes node representations accordingly. These representations have been used in several domains to obtain state of the art results in classification tasks where nodes were, for example, texts in a citation network or proteins in human tissues~\cite{velickovic2018graph}.
To our knowledge, we are the first to use Graph Attention Networks to model relations among social media users.

\section{Model}
\label{sec:model}

The model we present operates on annotated corpora made up of triples $(t, a, y)$, where $t$ is some user-generated text, $a$ is its author, and $y$ is a label classifying $t$. We address the task of predicting $y$ given $(t,a)$. Our focus is on user representations: We investigate how model predictions vary depending on how authors are represented.

\subsection{General Model Architecture}
\label{subsec:general_architecture}

\begin{figure*}[t]
\centering
\includegraphics[height=4.0cm]{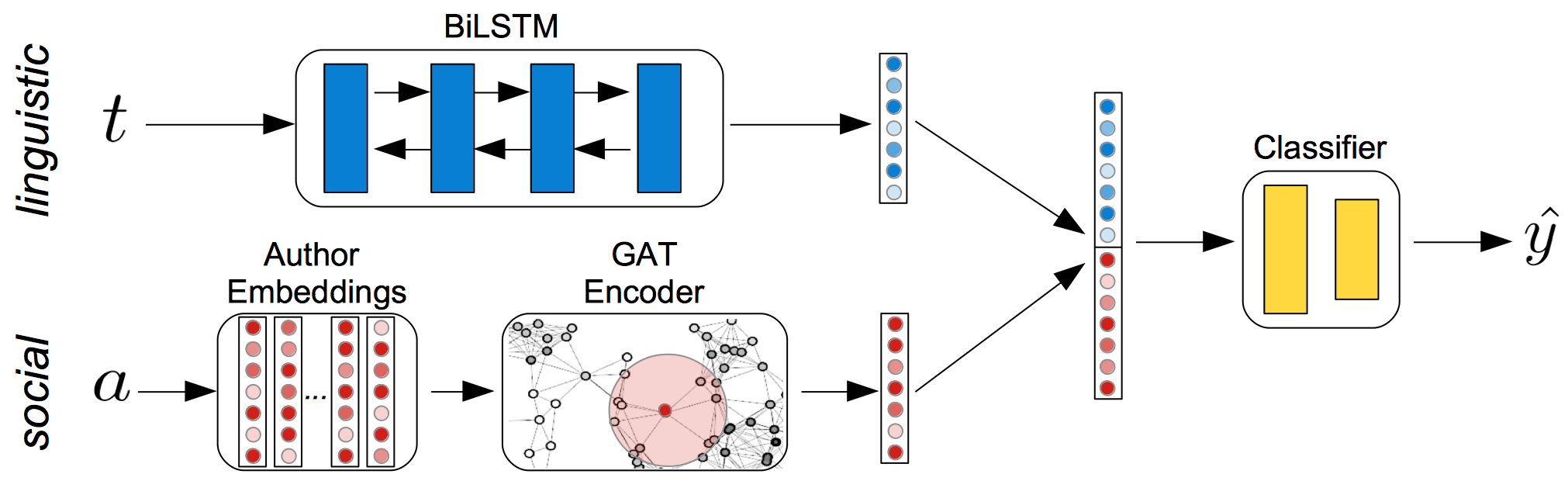}
\caption{General model: The linguistic module returns a compact representation of the input tweet \textit{t}. The social module takes as input the precomputed representation of the author \textit{a} and updates it using the GAT encoder. The output embeddings of the two modules are concatenated and fed into a classifier.}
\label{fig:model}
\end{figure*}

Our model consists of two modules, one encoding the linguistic information in $t$ and the other one modelling social information related to $a$. The general architecture is shown in Figure \ref{fig:model}. The output of the linguistic and social modules are vectors $l \in \mathbb{R}^{d}$ and $s \in \mathbb{R}^{d'}$, respectively. We adopt a standard \textit{late fusion} approach in which these two vectors are concatenated and passed through a two-layer classifier, consisting of a layer $W_1$ $\in \mathbb{R}^{d+d' \times c}$, where $c$ is a model parameter, and a layer $W_2$ $\in \mathbb{R}^{c \times o}$, where $o$ is the number of output classes. The final prediction is computed as follows, where $\sigma$ is a ReLU function \cite{nair2010rectified}:
\begin{align} \label{eq:general_model}
   \hat{y} = {\it  softmax}(W_2 \left( \sigma \left( W_1 \left( l \Vert s \right)\right)\right))
\end{align}

\subsection{Linguistic Module}
\label{subsec:linguistic_module}

The linguistic module is implemented using a recurrent neural network, concretely an \textsc{LSTM} \cite{hochreiter1997long}. Since LSTMs have become ubiquitous in NLP, we omit a detailed description of the inner workings of the model here and refer readers to \newcite{Tai2015,Tang2016,Barnes2017} for overviews. We use a bidirectional \textsc{LSTM} (\textsc{BiLSTM}) \cite{graves2012supervised}, whose final states are concatenated in order to obtain the representation of the input text. 

\subsection{Social Module}
\label{subsec:social_module}

The goal of the social module is to return author representations which encode homophily relations among users, 
i.e., which assign similar vectors to users who are socially related. 

We model social relations using graphs $G = (V,E)$, where $V$ is the set of nodes representing individuals  and $E$ the set of edges representing relations among them. We use $v_i \in V$ to refer to a node in the social graph, and $e_{ij} \in E$ to denote the edge connecting nodes $v_i$ and $v_j$. Finally, we use $N(v_i)$ to refer to $v_i$'s neighbours, i.e., all nodes which are directly connected to $v_i$.\footnote{Throughout the paper, we use the terms `individual', `author', `user' and `node' interchangeably.}

In contrast to most existing models, where user representations are static, our model uses an encoder which takes as input a pre-computed user vector, performs a dynamic exploration of its neighbours in the social graph, and updates the user representation given the 
relevance of its connections for a target task.
To pre-compute initial user representations we use Node2Vec \cite[N2V,][]{grover2016node2vec}.
Similarly to word2vec's SkipGram model \cite{mikolov2013word2vec}, for every node $v$, N2V implements a function $f:v\rightarrow\mathbb{R}^{d}$ which maps $v$ to a low-dimensional embedding of size $d$ that maximizes the probability of observing nodes belonging to  $S(v)$, i.e., the set of $n$ nodes encountered in the graph by taking $k$ random walks starting from $v$ (where $n$ and $k$ are parameters of the model).
No distinction is made among neighbours, thus ignoring the fact that different neighbours may have different importance depending on the task at hand.
Our model addresses this fundamental problem by leveraging Graph Attention Networks \cite[GATs,][]{velickovic2018graph}.


GATs extend the Graph Convolutional Networks proposed by \newcite{kipf2017semi}\footnote{For a detailed description of Graph Convolutional Networks see \url{https://tkipf.github.io/graph-convolutional-networks/}.} by introducing a self-attention mechanism \cite{bahdanau2015neural, parikh2016decomposable,vaswani2017attention} which is able to assign different relevance to different neighbouring nodes.
For a target node $v \in V$, an attention coefficient $e_{vu}$ is computed for every neighbouring node $u \in N(v)$ as:
\begin{align} \label{eq:attention_coefficient}
    e_{vu} = {att} \left (h_v \Vert h_u \right)
\end{align}
where $h_v$ and $h_u$ $\in \mathbb{R}^{d}$ are the vectors representing $v$ and $u$, $\Vert$ is concatenation,
and $att$ is a single-layer feed-forward neural network, parametrized by 
a weight matrix $W^a \in \mathbb{R}^{2d}$ with Leaky ReLU non-linearity \cite{maas2013rectifier}.
The attention coefficients for all the neighbours are then normalized using a softmax function.
%
%
Finally, the update of a node is computed as:
\begin{align} \label{eq:update_gat}
    h^{k+1}_v = \sigma \left( \sum_{u \in N(v)} \alpha_{vu}^k \left(W^k h^k_u + b^k\right)\right)
\end{align}
where  $W^k$ and $b^k$ are the layer-specific parameters of the model, 
$N(v)$ is the set of neighbours of the target node $v$, $h^{k+1}_v$ the updated node representation, $\sigma$ a ReLU function, $k$ the convolutional layer,\footnote{The number of layers defines the depth of the neighbourhood, hence $k$ layers encode neighbours lying $k$ hops away in the graph. Also, $v \in N(v)$, i.e.,  there exists a \textit{self-loop} $(v,v) \in E$ for each node $v$ which ensures that the initial representation of the node is considered during its update.} 
and $\alpha_{vu}$ the normalised attention coefficient, which defines how much neighbour $u$ should contribute to the update of $v$.

To stabilize the learning process, multiple attention mechanisms, or \textit{heads}, can be used. The number of heads is a hyperparameter of the model. Given $n$ heads, $n$ real-valued vectors $h^{k+1}_v \in \mathbb{R}^{d'}$ are computed 
and, subsequently, concatenated, thus obtaining a single embedding  $h^{k+1}_v \in \mathbb{R}^{n*d'}$. The resulting vector is then concatenated with the output of the linguistic module and fed into the classifier.

\section{Experimental Setup}
\label{sec:experiments}
 
\subsection{Alternative Models}
\label{subsec:alternative_models}

We compare the performance of our model against several 
competing models.
All the models except Frequency baseline and LING compute a user representation which is concatenated with the linguistic information present in a tweet and fed into the classifier.

\paragraph{Frequency Baseline}Labels are sampled according to their frequency in the whole dataset.

\paragraph{LING}
We use the linguistic module (LING) alone to assess the performance of the model when no social information is provided.

\paragraph{LING\texttt{+}random}
We implement a setting similar to 
\newcite{kolchinski2018representing}. Each individual is assigned a random embedding, which is updated during training. 
Our implementation differs from the \citeauthor{kolchinski2018representing}'s model in two aspects: They use GRUs \cite{cho2014learning} rather than LSTMs for the linguistic module, and their author vectors have size 15, while ours 200 for a fair comparison with the other 
models (see below).



\paragraph{LING\texttt{+}PV}
In line with \newcite{wallace2016modelling} and \newcite{hazarika2018cascade}, we compute a representation for each author by running Paragraph Vector \cite[PV,][]{le2014distributed} on the concatenation of all her previous tweets. Author representations are thus based on past linguistic usage.


\paragraph{LING\texttt{+}N2V}
While none of the previous models make use of the social graph, 
here we represent authors by means of the embeddings created with N2V \cite[as e.g.,][]{mishra2018author}. In contrast to our GAT-based model, the embeddings are computed without making any distinction among neighbours, and are not updated with respect to the task at hand.\footnote{We also experimented with updating author embeddings during training, but did not observe any difference in the results.} 

\subsection{Hyperparameter Search }
\label{subsec:hyperparameter_search}

For all the models and for each dataset, we perform grid hyperparameter search on the validation set using early stopping. 
For batch size, we explore values 4, 8, 16, 32, 64; for dropout, values 0.0, ..., 0.9; and for L2 regularisation, values 0, $1^{e-05}$, $1^{e-04}$.
For all the settings, we use Adam optimizer \cite{kingma2015adam} with learning rate of 0.001, $\beta_1$ = 0.9 and $\beta_2$ = 0.999. 
We run PV with the following hyperparameters: 30 epochs, minimum count of 5, vector size of 200. 
For N2V we use the default hyperparameters, except for vector size (200) and epochs (20). 
For the GAT encoder, we experiment with values 10, 15, 20, 25, 30, 50 for the size of the hidden layer; for the number of heads, we explore values 1, 2, 3, 4. We keep the number of hops equal to 1 and the alpha value for the Leaky ReLU of the attention heads equal to 0.2 across all the settings.\footnote{We implement PV using the Gensim library: \url{https://radimrehurek.com/gensim/models/doc2vec.html}. For N2V, we use the implementation at: \url{https://github.com/aditya-grover/node2vec}. For GAT, the implementation at: \url{https://github.com/Diego999/pyGAT}.}

Since our focus is on social information, we keep the hyperparameters of the linguistic module and the classifier fixed across all the settings. Namely, the BiLSTM has depth of 1, the hidden layer has 50 units, and uses 200-d GloVe embeddings pretrained on Twitter \cite{pennington2014glove}. 
For the classifier, we set the dimensionality of the non-linear layer to 50.


\subsection{Tasks and Datasets}
\label{subsec:tasks_and_datasets}

We test all the models on three Twitter datasets annotated for different tasks. For all datasets we tokenise and lowercase the text, and replace any URL, hashtag, and mention with placeholders. 

\paragraph{Sentiment Analysis}
We use the dataset in Task-4 of SemEval-2017 \cite{rosenthal2017semeval}, which includes 62k tweets labelled as \texttt{POSITIVE} (35.6\% of labels), \texttt{NEGATIVE} (18.8\%) and \texttt{NEUTRAL} (45.6\%).  
Tweets in the train set were collected between 2013 and 2015, while those in the test set in 2017.
Information for old tweets is difficult to recover:\footnote{Tweets can be deleted by users or administrators. This happens more often for old tweets.} To have a more balanced distribution, we shuffle the dataset and then split it into train (80\%), validation (10\%) and test (10\%). 


\paragraph{Stance Detection} 
We use the dataset released for Task-6 (Subtask A) of SemEval-2016 \cite{mohammad2016semeval}, which includes 4k tweets labelled as \texttt{FAVOR} (25.5\% of labels), \texttt{AGAINST} (50.6\%)  and \texttt{NEUTRAL} (23.9\%), with respect to five topics:~`Atheism', `Climate change is a real concern', `Feminist movement', `Hillary Clinton', `Legalization of abortion'. 
The dataset is split into train and test. We randomly extract 10\% of tweets in the train split and use them for validation.

\paragraph{Hate Speech Detection}
We employ the dataset introduced by \newcite{founta2018large}, from which we keep only tweets labelled as \texttt{NORMAL} (93.4\% of labels) and \texttt{HATEFUL} (6.6\%) for a total of 44k tweets.\footnote{The other labels in the dataset are \texttt{SPAM} and \texttt{ABUSIVE}.} The latter are tweets which denigrate a person or group based on social features (e.g., ethnicity). We split into train (80\%), validation (10\%) and test (10\%).



\subsection{Optimization Metrics}
\label{subsec:metrics}

We tune the models using different evaluation measures, according to the task at hand.
The rationale behind this choice is to use, whenever possible, established metrics per task. 

For Sentiment Analysis we use 
average recall, the same measure used for Task 4 of SemEval-2017 \cite{rosenthal2017semeval}, computed as:
\begin{align} \label{eq:ave_reca}
   AvgRec = \frac{1}{3} (R^{P} + R^{N} + R^{U} )
\end{align}
Where  $R^{P}$, $R^{N}$ and $R^{U}$ refer to recall of the \texttt{POSITIVE}, the \texttt{NEGATIVE}, and the \texttt{NEUTRAL} class, respectively. The measure has been shown to have several desirable properties, among which robustness to class imbalance \cite{sebastiani2015axiomatically}.

For Stance Detection, we use the average of the F-score of \texttt{FAVOR} and \texttt{AGAINST} classes:
\begin{align} \label{eq:ave_f_a}
   F_{avg} = \frac{F_{favor} + F_{against}}{2}
\end{align}
The measure, used for Task-6 (Subtask A) of SemEval-2016, is designed in such a way to optimize the performance of the model 
in the cases when an opinion toward the target entity is expressed, while it ignores the neutral class \cite{mohammad2016semeval}.  

Finally, for Hate Speech Detection, a more recent task, we could not identify an established metric. We thus use  
F-score for the target class \texttt{HATEFUL}, the minority class accounting for 6.6\% of the datapoints (see Section \ref{subsec:tasks_and_datasets}).

\subsection{Social Graph Construction}
\label{subsec:graphs}
In order to create the social graph, we initially retrieve, for each tweet, the \texttt{id} of its author using the Twitter API and scrape her timeline, i.e. her past tweets.\footnote{We access the API using the Python package Tweepy: \url{http://docs.tweepy.org/en/v3.5.0/}. The API returns a maximum of 3.2k tweets per user.}
We then create an independent social graph $G = (V,E)$ for each dataset. 
We define $V$ as the set of users authoring the tweets in the dataset, while for $E$ we follow \newcite{yang2017overcoming} and instantiate an unweighted and undirected edge between two users if one retweets the other. Information about retweets is available in users' timeline.
In order to make the graph more densely connected, we include external users not present in the dataset who have been retweeted by authors in the dataset at least 100 times. Information about the author of a tweet is not always available: In this case, we assign her an embedding computed as the centroid of the existing precomputed author representations. We note that in our datasets, authors with more than one tweet are very rare (6.6\% on average). 

Table \ref{tab:data} summarises the main statistics of the datasets and their respective graphs. The three social graphs have different number of nodes: the network of the Sentiment dataset is the largest ($\sim$62k nodes) while the Stance network is the smallest ($\sim$4k nodes). The number of edges and the density of the network (i.e., the ratio of existing connections over the number of potential connections) vary according to graph size, while the number of connected components is 1 for all the graphs: this means that there are no disconnected sub-graphs in the social networks. The most relevant aspect for which we find differences across the three graphs is the amount of homophily observed, which we define as the percentage of edges which connect users whose tweets have the same label. This value is higher for the Stance and Hate Speech social graphs than for the Sentiment one.
These figures indicate that, in our datasets, users expressing similar opinions about a topic (Stance) or using offensive language (Hate Speech) are more connected than those expressing the same sentiment in their tweets (Sentiment).

\begin{table}[h] 
\centering 
\begin{tabular}{@{}l|ccc@{}}
\toprule
\bf  & \bf  Sentiment & \bf Stance & \bf Hate \\ 
\midrule
\bf \# tweets      & 62,530 & 4,063 & 44,141 \\
\bf \% with author & 71.4\% & 71.7\% & 77.1\%\\
\bf \# nodes 	   & 50k      & 6.9k  	    & 25k \\
\bf \# edges 	   & 4.1m    & 258k  	 & 1.3m \\
\bf density 	   & 0.003    & 0.010  	 & 0.004 \\
\bf \# components 	   & 1    & 1  	 & 1 \\
\bf homophily  	   & 38\%    & 60\%  	 & 68\% \\
\bottomrule
\end{tabular}
\caption{Statistics for each dataset: Number of tweets; percentage of tweets for which we are able to retrieve information about the author; number of nodes; number of edges; density; number of connected components; and amount of homophily as percentage of connected authors whose tweets share the same label.}
\label{tab:data}
\end{table}


\section{Results}
\label{sec:results}

\begin{table}[t] 
\centering
\begin{tabular}{@{}lcll}
\toprule
\bf Model & \bf Sentiment & \bf Stance & \bf Hate \\
 \hline
  \small Frequency & \parbox{1.2cm}{\centering 0.332} & \parbox{1.2cm}{0.397} & \parbox{1.2cm}{0.057}  \\
\hline
 \small LING & \parbox{1.2cm}{\centering 0.676} & \parbox{1.2cm}{0.569} & \parbox{1.2cm}{0.624}  \\
%
 \small LING\texttt{+}random & \parbox{1.2cm}{\centering 0.657} & \parbox{1.2cm}{0.571} & \parbox{1.2cm}{0.600}  \\
%
 \small LING\texttt{+}PV & \parbox{1.2cm}{\centering 0.671} & \parbox{1.2cm}{0.601$^{\ast}$} & \parbox{1.2cm}{0.667$^{\ast}$}  \\
%
 \small LING\texttt{+}N2V & \parbox{1.2cm}{\centering 0.672} & \parbox{1.2cm}{0.629$^{\ast\diamond}$} & \parbox{1.2cm}{0.656$^{\ast}$}  \\
%
 \small LING\texttt{+}GAT & \parbox{1.2cm}{\centering 0.666} & \parbox{1.2cm}{0.640$^{\ast\diamond\dagger}$} & \parbox{1.2cm}{0.674$^{\ast\diamond\dagger}$}  \\
   \bottomrule
\end{tabular}
\caption{Results for all the models on the three datasets in our experiment. 
Marked with $^{\ast}$ are the results which significantly improve over LING and LING\texttt{+}random ($p\!<\!0.05$, also for the following results); 
$\diamond$ indicates a significant improvement over LING\texttt{+}PV; $\dagger$ a significant improvement over LING\texttt{+}N2V.\\
}
\label{tab:results}
\end{table}


We evaluate the performance of the models using the same metrics used for the optimization process (see Section \ref{subsec:metrics}). 
In Table \ref{tab:results} we report the results, that we compute as the average of ten runs with random parameter initialization.\footnote{The standard deviation for all models is small, in the range [0.003-0.02]. The full table, including the results for each class, is in the supplementary material.}
We use the unpaired Welch's $t$ test to check for 
statistically significant difference between models.

\paragraph{Tasks} 

The results show that social information helps improve the performance on Stance and Hate Speech detection, while it has no effect for Sentiment Analysis. 
While this result contrasts with the one reported by \newcite{yang2017overcoming}, who use a previous version of the Sentiment dataset \cite{rosenthal2015semeval},
it is not surprising given the analysis made in the previous section regarding the amount of homophily in the three social graphs: 
In our version of the data, sentiment is not as related to the social standing of individuals as stance and hatefulness are.
We reserve a deeper investigation of the 
impact of social information on the sentiment task to future work.

\paragraph{Models} 

LING\texttt{+}random never improves over LING: We believe this is due to
the fact that most of the authors have just one tweet, which hinders the possibility to learn at training time the representations of the users used at test time. 


We find that both PV and N2V user representations lead to an improvement over LING. N2V vectors are especially effective for the Stance detection task, where LING\texttt{+}N2V 
outperforms LING\texttt{+}PV, while for Hate Speech the performance of the two models is comparable (the difference between LING\texttt{+}PV and LING\texttt{+}N2V is not statistically significative due to the high variance of the LING\texttt{+}PV  results - see extended results table in the supplementary material).

Finally, our model outperforms any other model on both Stance and Hate Speech detection. This result confirms our initial hypothesis that a social attention mechanism which is able to assign different relevance to different neighbours allows for a more dynamic encoding of 
homophily relations in author embeddings and, in turn, leads to better results on the prediction tasks.\footnote{In preliminary experiments, Graph Convolutional Networks with no attention showed no improvements over the N2V baseline. The result is to be expected since, similarly to N2V, the model computes the representation of the target node without making any distinction among its neighbours.}

\section{Analysis}
\label{sec:analysis}

\begin{figure*}[t]
\hspace*{-.33cm}
\begin{minipage}{5.3cm}
\includegraphics[height=5cm]{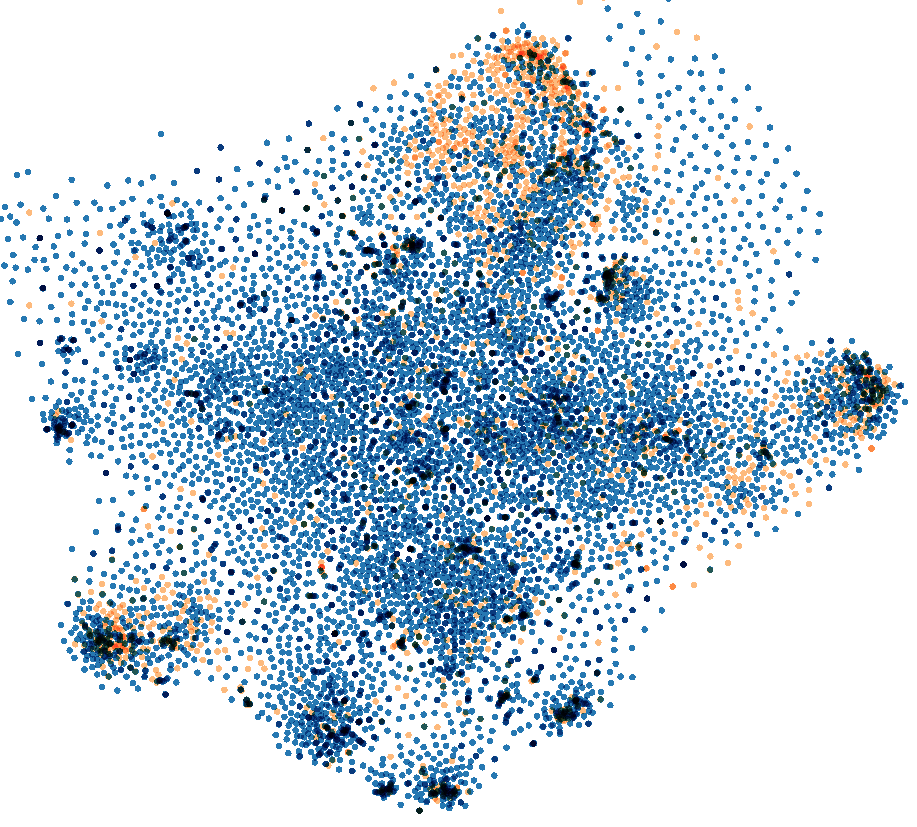}
\end{minipage} \ \ 
\begin{minipage}{4.9cm}
\includegraphics[height=4.9cm]{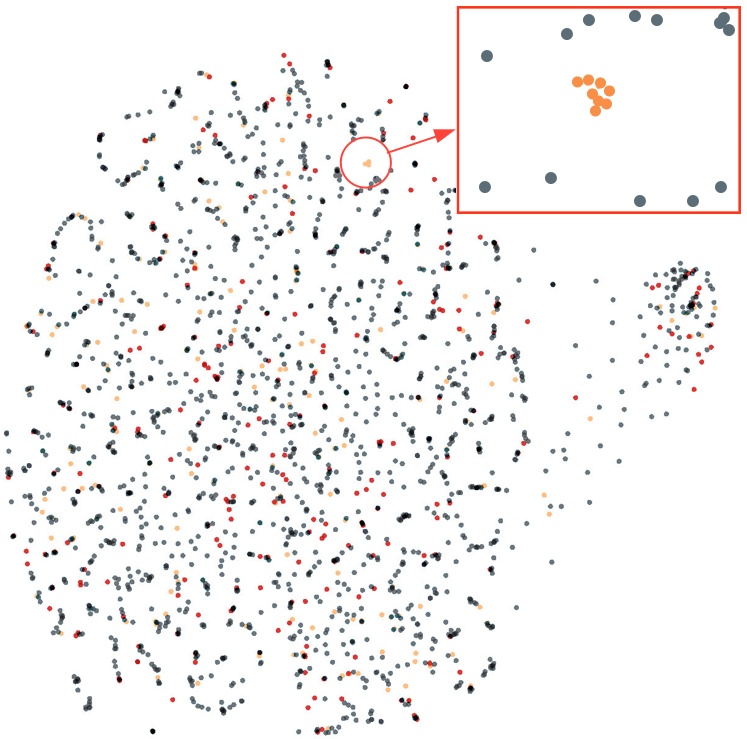}
\end{minipage}
\begin{minipage}{5.1cm}
\includegraphics[height=5cm]{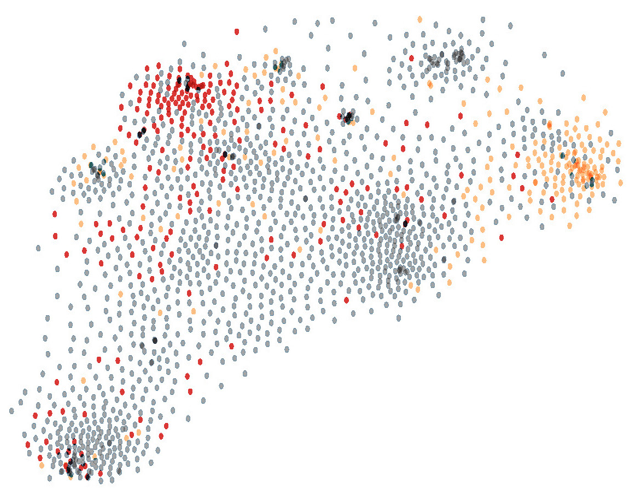}
\end{minipage}
\caption{\textbf{Left}: PV user representations for the Hate Speech dataset. Orange dots are authors of \texttt{HATEFUL} tweets, blue dots of \texttt{NORMAL} tweets. \textbf{Center} and \textbf{Right}: PV and N2V user representations, respectively, for the Stance dataset. In both plots, orange dots are authors of tweets \texttt{AGAINST} atheism, red dots authors in \texttt{FAVOR} of `climate change is a real concern'. 
All other users are represented as grey dots.}
\label{fig:vector_spaces}
\end{figure*}

%


\begin{table}[t!] \centering \small 
\begin{tabular}{@{}l@{\ }p{7cm}@{}}
\toprule
(1) & \textit{@user:~Yurtle the Turtle needs to be slapped}\\
      & \textit{with a f***ing chair..many times!} \hfill \texttt{HATEFUL}\\\midrule
(2)  & \textit{You stay the same through the ages\ldots Your love never changes\ldots Your love never fails}\\
       & \hfill  \texttt{AGAINST atheism}\\\midrule
(3) & \textit{Why are Tumblr feminists so territorial?}\\
      & \textit{Pro-lifers can't voice their opinions without}\\
      & \textit{being attacked} \hfill \texttt{AGAINST abortion}\\ \midrule
(4) & \textit{@user No, just pointed out how idiotic your}\\
      & \textit{statement was} \hfill \texttt{HATEFUL}\\
\bottomrule
\end{tabular}
\caption{Examples from Stance (2, 3) and Hate Speech (1, 4) datasets, and their gold label.
}
\label{tab:examples}
\end{table}

We analyse in more detail the strengths and weaknesses of the best models for the tasks where social information proved useful.

\subsection{Paragraph Vector}
\label{subsec:paragraph_vector}

Figure~\ref{fig:vector_spaces} (left) shows the user representations created with PV for the Hate Speech dataset.\footnote{Plots are created using the Tensorflow Projector available at: \url{https://projector.tensorflow.org}.} 
The plot shows that users form sub-communities, with authors of hateful tweets (orange dots) mainly clustering at the top of the plot. The similarity between these individuals derives from their consistent use of strongly offensive words towards others over their posting history. This suggests that representing speakers in terms of their past linguistic usage can, to some extent, capture certain homophily relations. 
For example,  tweet (1) in Table~\ref{tab:examples} is incorrectly labelled as \texttt{NORMAL} by the LING model.\footnote{Note that {\em `f***ing'} is often used with positive emphasis and is thus not a reliable clue for hate speech.} By leveraging the PV author representation (which, given previous posting behaviour, is highly similar to authors of hateful tweets) the LING\texttt{+}PV model yields the right prediction in this case.

For Stance detection, which arguably is a less lexically determined task \cite{mohammad2016semeval}, PV user representations are less effective. This is illustrated in Figure~\ref{fig:vector_spaces} (center), where no clear clusters are visible. Still, PV vectors capture some meaningful relations, such as a small, close-knit cluster of users against atheism (see zoom in the figure), who tweet mostly about Islam.


\subsection{Node2Vec}
\label{subsec:node2vec}

User representations created by exploiting the social network of individuals are more robust across datasets. For Hate Speech, the user representations computed with PV and N2V are very similar.\footnote{The plot showing  N2V user representations for the Hate Speech dataset is in the supplementary material.} However, for the Stance dataset, N2V user representations are more informative. This is readily apparent when comparing the plots in the center and right of 
Figure~\ref{fig:vector_spaces}: Users who were scattered when represented with PV now form community-related clusters, which leads to better predictions.  For example, tweet (2) in Table~\ref{tab:examples} is authored by a user who is socially connected to other users who tweet against atheism (the orange cluster in the right-hand side plot of Figure~\ref{fig:vector_spaces}).  The LING\texttt{+}N2V model is able to exploit this information and make the right prediction, while the tweet is incorrectly classified by LING and LING\texttt{+}PV, which do not take into account the author's social standing.

N2V, however, is not effective for users connected to multiple sub-communities, because by definition the model will conflate this information into a fixed vector located between clusters in the social space.\footnote{This effect has some parallels with how word2vec derives representations for polysemous words.}
For instance, the author of tweet (1) is connected to both users who post hateful tweets and users whose posts are not hateful. In the N2V user space, the ten closest neighbours of this author are equally divided between these two groups. In this case, the social network information captured by N2V is not informative enough and, as a result, the tweet ends up being wrongly labeled as \texttt{NORMAL} by the LING\texttt{+}N2V model, i.e., there is no improvement over LING.


\subsection{Graph Attention Network}
\label{subsec:gat}

As hypothesised, the GAT model allows us to address the shortcoming of N2V we have described above. When creating a representation for the author of tweet (1), the GAT encoder identifies the connection to one of the authors of a hateful tweet as the most relevant for the task at hand, and assigns it the highest value. The user vector is updated accordingly, which results in the LING\texttt{+}GAT model correctly predicting the \texttt{HATEFUL} label. 

\begin{figure}[t]\centering
\includegraphics[width=\columnwidth]{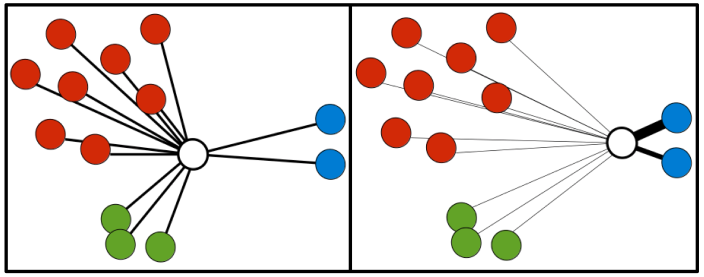}
\caption{{\bf Left:} 
Author of tweet (3) (white node) has many connections with users tweeting in favour of feminism (red dots), fewer with authors tweeting in favour of Clinton (green dots) and against legalization of abortion (blue dots) (for simplicity, only some connections are shown). Recall that no label is available for nodes (users), and that their color is based on the label of the tweet they posted. Proximity of dots in the space reflects vector similarity. 
{\bf Right:} the GAT encoder assigns higher values to connections with the relevant neighbours (0.54 and 0.14; all other connections have values $\leq$ 0.02; thickness of the edges is proportional to their values) and updates the target author vector to make it proximal to them in social space.
}
\label{fig:gat}
\end{figure}

 This dynamic exploration of the social connections has the capacity to highlight homophily relations that are less prominent in the social network of a user, but more relevant in a given context. This is illustrated by how the models deal with tweet (3)  in Table~\ref{tab:examples}, which expresses a negative stance towards legalisation of abortion, and is incorrectly classified by LING and LING\texttt{+}PV.  
The social graph contributes rich information about its author, who is connected to many users (46 overall). Most of them are authors who tweet in favour of feminism: 
The N2V model effectively captures this information, as the representation of the target user is close to these authors in the vector space. 
Consequently, by simply focusing on the majority of the neighbourhood, the LING\texttt{+}N2V model misclassifies the tweet (i.e., it infers \texttt{FAVOR} for tweet (3) on the legalisation of abortion from a social environment that mostly expresses stances in favour of feminism). 
However, the information contributed by the majority of neighbours is not the most relevant in this case. In contrast, 
%
the GAT encoder identifies the connections with two users who tweet against legalisation of abortion as the most relevant ones, 
and updates the author representation in such a way to increase the similarity with them -- see Figure \ref{fig:gat} for an illustration of this dynamic process -- which leads the LING\texttt{+}GAT model to make the right prediction. 

Interestingly, GAT is able to recognise when the initial N2V representation already encodes the necessary information. For example, the LING\texttt{+}N2V model correctly classifies tweet (4)  in Table~\ref{tab:examples}, as the N2V vector of its author is close in social space to that of other users who post hateful tweets (7 out of 10 closest neighbours). In this case, the LING\texttt{+}GAT model assigns the highest value to the self-loop connection, thus avoiding to modify a representation which is already well tuned for the task.

Our error analysis reveals that there are two main factors which affect the performance of the GAT model. 
One is the size of the neighbourhood: As the size increases, the normalised attention values tend to be very small and equally distributed, which makes the model incapable of identifying relevant connections.
The second is related to the fact that a substantial number of users ($\sim$800 for Stance and $\sim$2.4k for Hate Speech) are not connected to the relevant sub-community. This means that in the case of Stance, for example, a user is not connected to any other individual expressing the same stance towards a certain topic. 
While external nodes in the graph (see Section \ref{subsec:graphs}) help to alleviate the problem by allowing the information to propagate through the graph, this lack of connections is detrimental to GAT.

\section{Conclusion}
\label{sec:conc}

In this work, we investigated representations for users in social media and their usefulness in downstream NLP tasks.
We introduced a model that captures the fact that not all the social connections of an individual are equally relevant in different communicative situations. The model dynamically explores the connections of a user, identifies the ones that are more relevant for a specific task, and computes her representation accordingly.
We showed that, when social information is proved useful, the dynamic representations computed by our model better encode homophily relations compared to the static representations obtained with other models.
In contrast to most models proposed in the literature so far, which are tested on one single task, we applied our model to three tasks, comparing its performance against several competing models.
Finally, we performed an extended analysis of the performance of all the models that effectively encode author information, highlighting strengths and weaknesses of each model.

In future work, we plan to perform a deeper investigation of cases in which social information does not prove beneficial, and to assess the ability of our model to dynamically update the representation of the \textit{same} author in different contexts, a task that, due to the nature of the data, was not possible in present work.


\section{Acknowledgements}
\label{sec:acknowledgements}

This research was partially funded by the Netherlands Organisation for Scientific Research (NWO) under VIDI grants \textit{Asymmetry in Conversation} (276-89-008) and \textit{Semantic Parsing in the Wild} (639-022-518). We are grateful to the Integrated Research Training Group of the Collaborative Research Center SFB 732 for generously funding a research visit by Marco Del Tredici to the University of Stuttgart that fostered some of the ideas in this work. Finally, we kindly thank Mario Giulianelli and Jeremy Barnes for their valuable contribution to earlier versions of the models.


\bibliography{emnlp-ijcnlp-2019}
\bibliographystyle{acl_natbib}

\appendix
\section{Supplemental Material}
\label{sec:supplemental}

\subsection{Extended Results Table}


Tables \ref{tab:sentiment}, \ref{tab:stance} and \ref{tab:hate} in next page include, for each task, the results for LING, LING\texttt{+}PV, LING\texttt{+}N2V and LING\texttt{+}GAT as reported in the main paper, together with the Standard Deviation values computed on the ten runs of each model. 
Additionally, the precision, recall and F-score for each class are reported. 


\begin{table*}[h] 
\centering
\begin{tabular}{lc|ccc|ccc|ccc}
\toprule
\bf Model & \bf Av. Rec. &  & \bf Negative &  &  & \bf Neutral &   &   & \bf Positive  &  \\
 \hline
 &  & \bf P & \bf R & \bf F1 & \bf P & \bf R & \bf F1 & \bf P & \bf R & \bf F1 \\
 \hline
 \small \parbox[][1.1cm][c]{1.7cm}{LING} & \parbox{0.8cm}{\centering 0.676 \\ \scriptsize{$\pm$0.005} } & \parbox{0.8cm}{\centering 0.585} & \parbox{0.8cm}{\centering 0.656} & \parbox{0.8cm}{\centering 0.618} & \parbox{0.8cm}{\centering 0.684} & \parbox{0.8cm}{\centering 0.678} & \parbox{0.8cm}{\centering 0.680} & \parbox{0.8cm}{\centering 0.737} & \parbox{0.8cm}{\centering 0.694} & \parbox{0.8cm}{\centering 0.712}\\

 \small \parbox[][1.1cm][c]{1.7cm}{LING+PV} & \parbox{0.8cm}{\centering 0.671 \\ \scriptsize{$\pm$0.004} } & \parbox{0.8cm}{\centering 0.584} & \parbox{0.8cm}{\centering 0.639} & \parbox{0.8cm}{\centering 0.609} & \parbox{0.8cm}{\centering 0.677} & \parbox{0.8cm}{\centering 0.679} & \parbox{0.8cm}{\centering 0.678} & \parbox{0.8cm}{\centering 0.734} & \parbox{0.8cm}{\centering 0.693} & \parbox{0.8cm}{\centering 0.713}\\

 \small \parbox[][1.1cm][c]{1.7cm}{LING+N2V} & \parbox{0.8cm}{\centering 0.672 \\ \scriptsize{$\pm$0.004} } & \parbox{0.8cm}{\centering 0.584} & \parbox{0.8cm}{\centering 0.639} & \parbox{0.8cm}{\centering 0.609} & \parbox{0.8cm}{\centering 0.681} & \parbox{0.8cm}{\centering 0.679} & \parbox{0.8cm}{\centering 0.680} & \parbox{0.8cm}{\centering 0.734} & \parbox{0.8cm}{\centering 0.699} & \parbox{0.8cm}{\centering 0.715}\\

 \small \parbox[][1.1cm][c]{1.7cm}{LING+GAT} & \parbox{0.8cm}{\centering 0.666 \\ \scriptsize{$\pm$0.01} } & \parbox{0.8cm}{\centering 0.599} & \parbox{0.8cm}{\centering 0.597} & \parbox{0.8cm}{\centering 0.596} & \parbox{0.8cm}{\centering 0.666} & \parbox{0.8cm}{\centering 0.691} & \parbox{0.8cm}{\centering 0.677} & \parbox{0.8cm}{\centering 0.730} & \parbox{0.8cm}{\centering 0.691} & \parbox{0.8cm}{\centering 0.709}\\
 
\bottomrule
\end{tabular}

\caption{\textbf{Sentiment Analysis}: Average Recall across the three classes, plus precision, recall and F1 per class.}
\label{tab:sentiment}
\end{table*}


\begin{table*}[h] 
\centering
\begin{tabular}{lc|ccc|ccc|ccc}
\toprule
\bf Model & \bf Av. Ag./Fa. &  & \bf Against &  &  & \bf Neutral &   &   & \bf Favor  &  \\
 \hline
 &  & \bf P & \bf R & \bf F1 & \bf P & \bf R & \bf F1 & \bf P & \bf R & \bf F1 \\
 \hline
 \small \parbox[][1.1cm][c]{1.7cm}{LING} & \parbox{0.8cm}{\centering 0.569 \\ \scriptsize{$\pm$0.01} } & \parbox{0.8cm}{\centering 0.730} & \parbox{0.8cm}{\centering 0.625} & \parbox{0.8cm}{\centering 0.672} & \parbox{0.8cm}{\centering 0.355} & \parbox{0.8cm}{\centering 0.462} & \parbox{0.8cm}{\centering 0.399} & \parbox{0.8cm}{\centering 0.446} & \parbox{0.8cm}{\centering 0.490} & \parbox{0.8cm}{\centering 0.466}\\

 \small \parbox[][1.1cm][c]{1.7cm}{LING+PV} & \parbox{0.8cm}{\centering 0.601$^{\ast}$ \\ \scriptsize{$\pm$0.02} } & \parbox{0.8cm}{\centering 0.739} & \parbox{0.8cm}{\centering 0.673} & \parbox{0.8cm}{\centering 0.701} & \parbox{0.8cm}{\centering 0.353} & \parbox{0.8cm}{\centering 0.380} & \parbox{0.8cm}{\centering 0.362} & \parbox{0.8cm}{\centering 0.479} & \parbox{0.8cm}{\centering 0.536} & \parbox{0.8cm}{\centering 0.501}\\

 \small \parbox[][1.1cm][c]{1.7cm}{LING+N2V} & \parbox{0.8cm}{\centering 0.629$^{\ast\diamond}$ \\ \scriptsize{$\pm$0.01} } & \parbox{0.8cm}{\centering 0.761} & \parbox{0.8cm}{\centering 0.697} & \parbox{0.8cm}{\centering 0.727} & \parbox{0.8cm}{\centering 0.380} & \parbox{0.8cm}{\centering 0.369} & \parbox{0.8cm}{\centering 0.370} & \parbox{0.8cm}{\centering 0.488} & \parbox{0.8cm}{\centering 0.588} & \parbox{0.8cm}{\centering 0.531}\\

 \small \parbox[][1.1cm][c]{1.7cm}{LING+GAT} & \parbox{0.8cm}{\centering 0.640$^{\ast\diamond\dagger}$ \\ \scriptsize{$\pm$0.01} } & \parbox{0.8cm}{\centering 0.749} & \parbox{0.8cm}{\centering 0.725} & \parbox{0.8cm}{\centering 0.734} & \parbox{0.8cm}{\centering 0.380} & \parbox{0.8cm}{\centering 0.316} & \parbox{0.8cm}{\centering 0.330} & \parbox{0.8cm}{\centering 0.507} & \parbox{0.8cm}{\centering 0.600} & \parbox{0.8cm}{\centering 0.545}\\
 
\bottomrule

\end{tabular}

\caption{\textbf{Stance Detection}: Average F1 of the Against and Favor classes, plus precision, recall and F1 per class.}
\label{tab:stance}
\end{table*}




\begin{table*}[h] 

\centering
\begin{tabular}{lc|ccc|ccc}
\toprule
\bf Model & \bf F1 Hateful &  & \bf Normal &  &  & \bf Hateful &   \\
 \hline
 &  & \bf P & \bf R & \bf F1 & \bf P & \bf R & \bf F1  \\
 \hline
\small \parbox[][1.1cm][c]{1.7cm}{LING} & \parbox{0.8cm}{\centering 0.624 \\ \scriptsize{$\pm$0.01} } & \parbox{0.8cm}{\centering 0.968} & \parbox{0.8cm}{\centering 0.989} & \parbox{0.8cm}{\centering 0.978} & \parbox{0.8cm}{\centering 0.773} & \parbox{0.8cm}{\centering 0.526} & \parbox{0.8cm}{\centering 0.624} \\

 \small \parbox[][1.1cm][c]{1.7cm}{LING+PV} & \parbox{0.8cm}{\centering 0.667$^{\ast}$ \\ \scriptsize{$\pm$0.02} } & \parbox{0.8cm}{\centering 0.974} & \parbox{0.8cm}{\centering 0.983} & \parbox{0.8cm}{\centering 0.979} & \parbox{0.8cm}{\centering 0.730} & \parbox{0.8cm}{\centering 0.621} & \parbox{0.8cm}{\centering 0.667} \\

 \small \parbox[][1.1cm][c]{1.7cm}{LING+N2V} & \parbox{0.8cm}{\centering 0.656$^{\ast}$ \\ \scriptsize{$\pm$0.008} } & \parbox{0.8cm}{\centering 0.972} & \parbox{0.8cm}{\centering 0.986} & \parbox{0.8cm}{\centering 0.979} & \parbox{0.8cm}{\centering 0.742} & \parbox{0.8cm}{\centering 0.589} & \parbox{0.8cm}{\centering 0.656}\\

 \small \parbox[][1.1cm][c]{1.7cm}{LING+GAT} & \parbox{0.8cm}{\centering 0.674$^{\ast\diamond\dagger}$ \\ \scriptsize{$\pm$0.005} } & \parbox{0.8cm}{\centering 0.973} & \parbox{0.8cm}{\centering 0.989} & \parbox{0.8cm}{\centering 0.980} & \parbox{0.8cm}{\centering 0.765} & \parbox{0.8cm}{\centering 0.605} & \parbox{0.8cm}{\centering 0.674} \\
 
\bottomrule
\end{tabular}

\caption{\textbf{Hate Speech Detection}: F1 for the Hateful class, plus precision, recall and F1 per class.}

\label{tab:hate}
\end{table*}

\clearpage


\vfill\eject
\section{Additional Plots}

The plot representing the N2V representations for the Hate Speech dataset. Orange dots are authors of tweets labelled as  \texttt{HATEFUL}, blue dots of \texttt{NORMAL} tweets.

\begin{figure}[h]\centering
\includegraphics[width=\columnwidth]{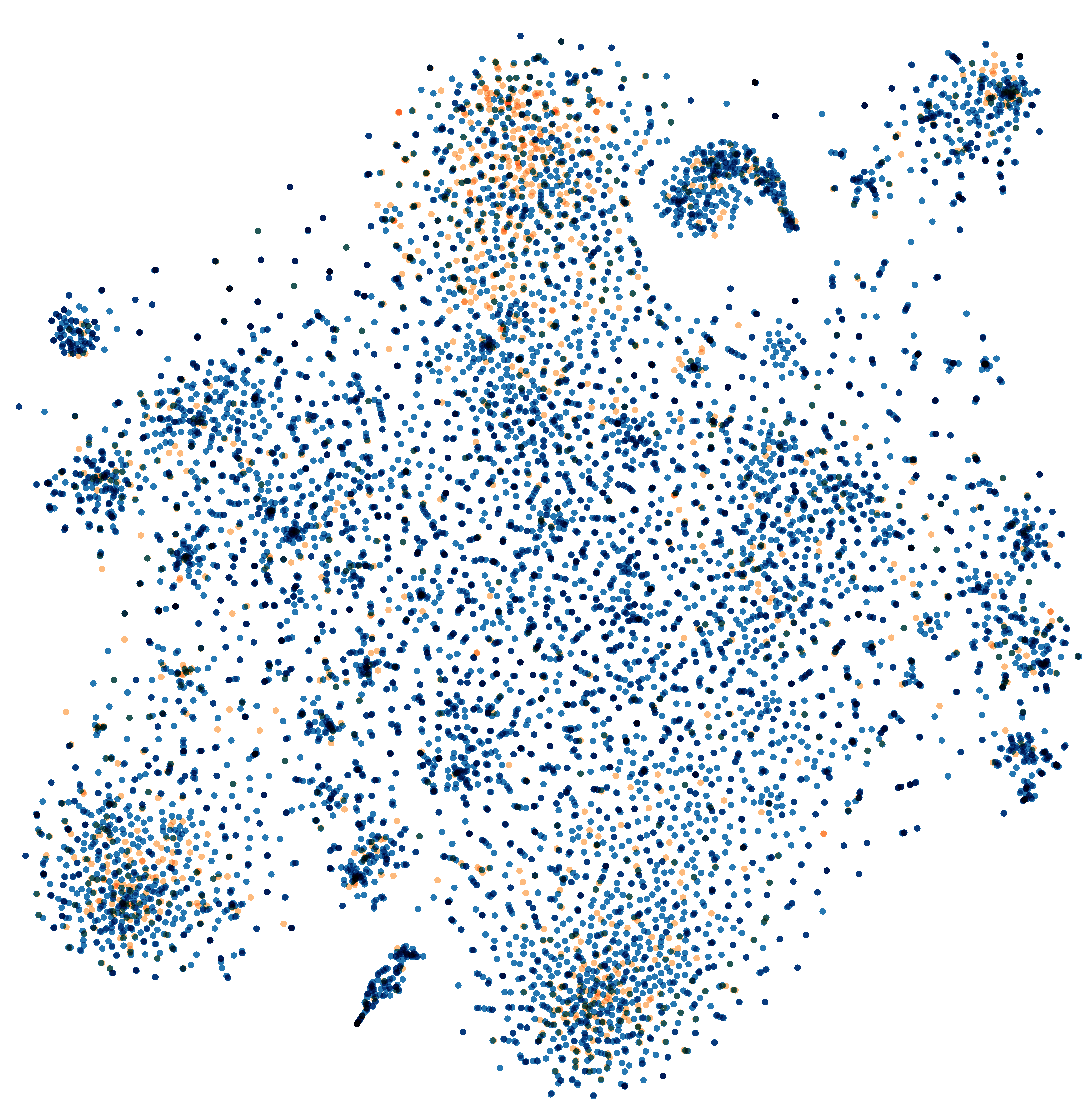}
\end{figure}


Below, the same plots in Figure (2) in the main paper showed in a larger size.
%
%

\begin{figure}[h]
\begin{minipage}{7cm}
\includegraphics[width=\columnwidth]{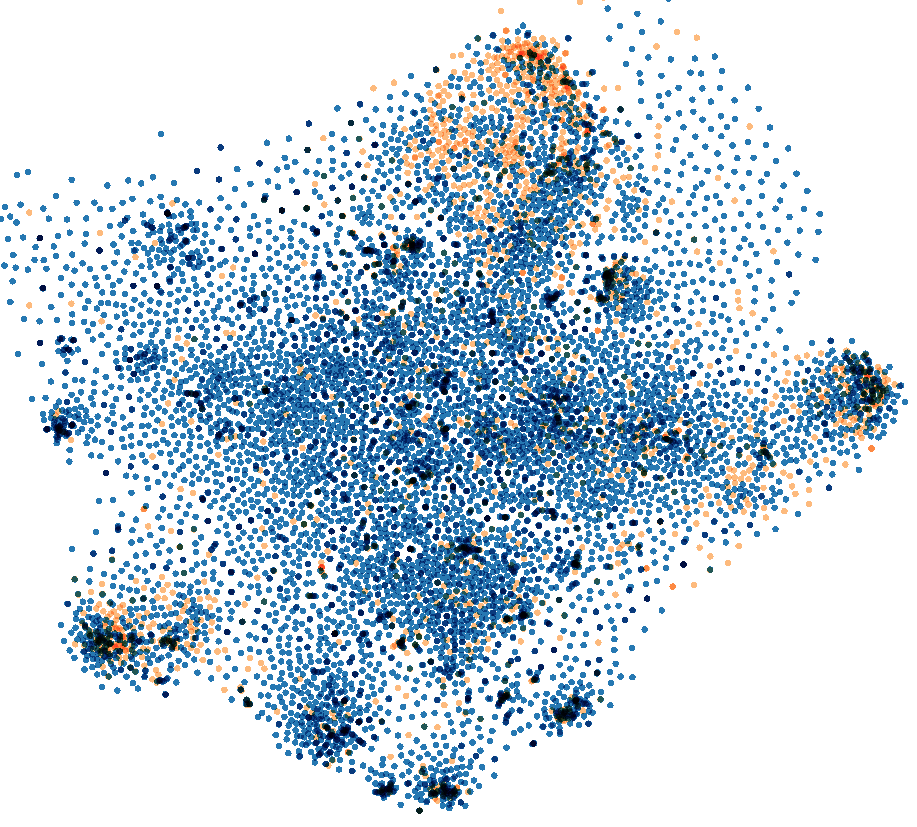}\\
\caption{PV user representations for the Hate Speech dataset. Orange dots are authors of \texttt{HATEFUL} tweets, blue dots of \texttt{NORMAL} tweets.}
\end{minipage} 
\end{figure}

\begin{figure}[h]
\begin{minipage}{7cm}
\includegraphics[width=\columnwidth]{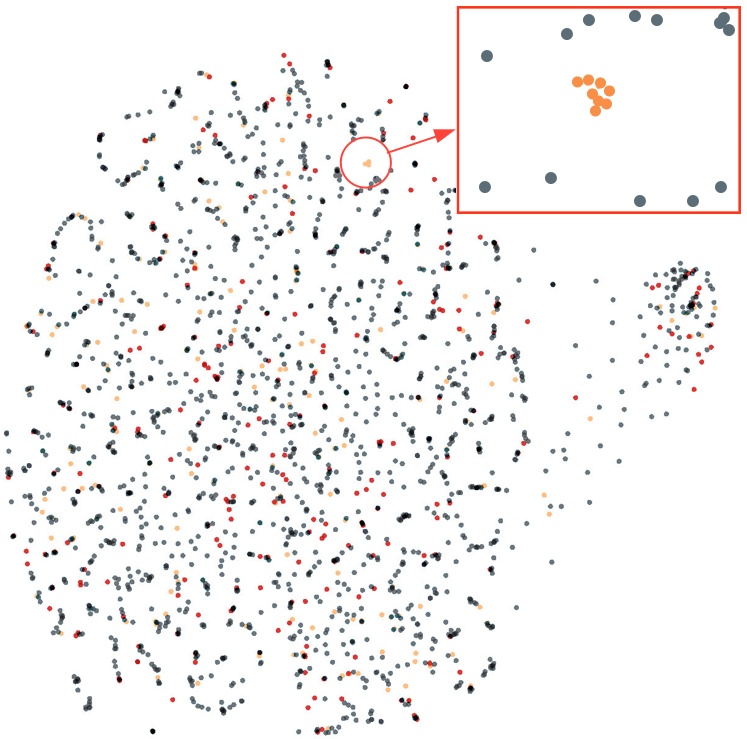}\\
\caption{PV user representations for the Stance dataset. Orange dots are authors of tweets \texttt{AGAINST} atheism, red dots authors in \texttt{FAVOR} of `climate change is a real concern'. All other users are represented as grey dots.}
\end{minipage} 
\end{figure}


\vspace*{0.5cm}
\begin{figure}[h]

\hspace*{10cm}

\begin{minipage}{7.5cm}
\includegraphics[width=\columnwidth]{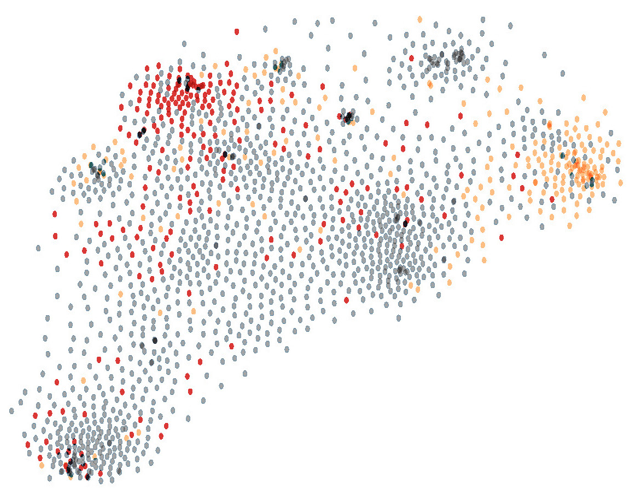}\\
\caption{N2V user representations for the Stance dataset. Orange dots are authors of tweets \texttt{AGAINST} atheism, red dots authors in \texttt{FAVOR} of `climate change is a real concern'. All other users are represented as grey dots.}
\end{minipage} 
\end{figure}

\end{document}